\documentclass{article} 
\usepackage{iclr2021_conference,times}
\iclrfinalcopy{\iclrfinaltrue}

\usepackage{amsmath,amsfonts,bm}

\def\eqref#1{equation~\ref{#1}}

\def\1{\bm{1}}

\def\rvx{{\mathbf{x}}}

\def\rvx{{\mathbf{x}}}

\DeclareMathAlphabet{\mathsfit}{\encodingdefault}{\sfdefault}{m}{sl}
\SetMathAlphabet{\mathsfit}{bold}{\encodingdefault}{\sfdefault}{bx}{n}

\newcommand{\E}{\mathbb{E}}

\usepackage{float}
\usepackage{hyperref}
\usepackage{url}
\usepackage{algorithm} 
\usepackage[noend]{algpseudocode} 
\usepackage{multirow}

\usepackage{booktabs} 
\usepackage{subcaption}

\usepackage{graphicx}

\title{EBMs Trained with Maximum Likelihood are Generator Models Trained with a Self-adverserial Loss}

\author{
  Zhisheng Xiao \thanks{equal contribution} \\
  Computational and Applied Mathematics\\
  University of Chicago\\
  Chicago, IL, 60637\\
  \texttt{zxiao@uchicago.edu} \\
   \And
  Qing Yan \footnotemark[1]\\
  Department of Statistics\\
  University of Chicago\\
  Chicago, IL, 60637\\
  \texttt{yanq@uchicago.edu} \\
    \And
  Yali Amit \\
  Department of Statistics\\
  University of Chicago\\
  Chicago, IL, 60637\\
  \texttt{amit@marx.uchicago.edu} \\
}

\begin{document}

\maketitle

\begin{abstract}
Maximum likelihood estimation is widely used in training Energy-based models (EBMs). Training requires samples from an unnormalized distribution, which is usually intractable, and in practice, these are obtained by MCMC algorithms such as Langevin dynamics. However, since MCMC in high-dimensional space converges extremely slowly, the current understanding of  maximum likelihood training, which assumes approximate samples from the model can be drawn, is problematic \citep{nijkamp2019learning}. In this paper, we try to understand this training procedure by replacing Langevin dynamics with deterministic solutions of the associated gradient descent ODE. Doing so allows us to study the density induced by the dynamics (if the dynamics are invertible), and connect with GANs by treating the dynamics as generator models, the initial values as latent variables and the loss as optimizing a critic defined by the very same energy that determines the generator through its gradient. Hence the term - self-adversarial loss. We show that reintroducing the noise in the dynamics does not lead to a qualitative change in the behavior, and merely reduces the quality of the generator.  We thus show that EBM training is effectively a self-adversarial procedure rather than maximum likelihood estimation. 
\end{abstract}

\section{Introduction}
Energy-based models (EBMs) are likelihood-based generative models that model the unnormalized data density by assigning low energy to high-probability regions in the data space. Recently, by using neural network as the energy functions, deep EBMs \citep{xie2016theory,du2019implicit} are able to model complex data such as natural images \citep{xiao2021vaebm, gao2021learning}, 3D shapes \citep{9294054} and texts \citep{deng2020residual}. There are a variety of ways to train EBMs, including minimizing the KL-divergence \citep{du2019implicit} or general F-divergence  \citep{yu2020training}, score matching \citep{li2019learning} and contrastive estimation \citep{gutmann2010noise,gao2020flow}. Among them, the KL divergence minimization (equivalent to maximum likelihood estimation) is most widely used. 
\subsection{Maximum likelihood training of EBMs}\label{mle intro}
To train an EBM of the form $p_{\theta}(\rvx)=\exp \left(-E_{\theta}(\rvx)\right) / Z_{\theta}$, where $E_{\theta}(\rvx)$ is the energy function with parameters $\theta$ and $Z_{\theta}=\int_{\rvx} \exp \left(-E_{\theta}(\rvx)\right) d\rvx$ is the normalizing constant, we can take the derivative of the negative log likelihood function $L(\theta) = \E_{\rvx \sim p_{d}(\rvx)}\left[-\log p_{\theta}(\rvx)\right]$ w.r.t to the model parameter $\theta$ \citep{woodford2006notes}:
\begin{align}
\label{derivative}
    \partial_\theta L(\theta) = \E_{\rvx \sim p_{d}\left(\rvx\right)}\left[ \partial_\theta E_{\theta}\left(\rvx\right)\right] - \E_{\rvx \sim p_{\theta}\left(\rvx\right)}\left[\partial_{\theta} E_{\theta}\left(\rvx\right)\right]
\end{align}
and minimize $L(\theta)$ by gradient descent. The second expectation in (\ref{derivative}) can be empirically estimated by samples drawn from the model $p_{\theta}(\rvx)$ itself. However, sampling from $p_{\theta}(\rvx)$ is intractable and samples are usually drawn using MCMC. A commonly used MCMC algorithm is the Langevin dynamics (LD)~\citep{neal1993probabilistic}. Given an initial sample $\rvx_0$, Langevin dynamics solves the SDE 
\begin{equation}\label{LD_cont}
d \rvx_t = -\frac{1}{2}\nabla_{\rvx}E_\theta(\rvx_t)dt + d \mathbf{w}_t,
\end{equation}
where $\mathbf{w}_t$ is Brownian motion.
The discretized version, using the simplest Euler approximation
yields:
 \begin{align}\label{LD formula}
    \rvx_{t+1}=\rvx_{t}-\frac{\eta}{2} \nabla_{\rvx} E_\theta(\rvx_t)+\sqrt{\eta} \mathbf{\omega}_t,
\end{align} where $\mathbf{\omega}_t \sim \mathcal{N}(0, \mathbf{I})$ and $\eta$ is the step-size. Theoretically, we need to run the discretized LD with infinitely many steps and diminishing step sizes to obtain true samples. However, in practice, we usually run LD for finite number of steps with a fixed step size. After training, samples are obtained by running the same Langevin dynamics, typically with the same number of steps. 

\subsection{Alternative understanding of maximum likelihood training}
Although the maximum likelihood training scheme is simple and intuitively appealing, we might still have not fully understood its mechanism. Since the convergence of MCMC is extremely difficult when the energy function is complicated, we cannot easily overlook the gap between running the LD in practice (usually called short-run LD) and truly obtaining samples from $p_{\theta}(\rvx)$. Indeed, some interesting observations are made from training the EBMs through maximum likelihood. Firstly, in practice the noise scale of LD is usually much smaller than the correct one in (\ref{LD formula}), which makes the LD similar to gradient descent \citep{du2019implicit}. Secondly, unless the shape of the energy function is carefully modified by introducing a base distribution as done in \citet{nijkamp2020learning, xiao2020exponential}, LD usually does not mix, i.e., samples obtained by running longer LD get trapped in different local modes instead of traversing between modes. Probably as a consequence, the initial points $\rvx_0$ contain information about the final outcome, and therefore short-run LD is observed to be capable of reconstructing the data and interpolating different samples \citep{nijkamp2019learning}. In addition, sometimes while we can obtain good samples by running short-run LD, the density of the EBMs can be drastically different from the true data densities (e.g., Figure 2 of \citet{gao2021learning}). These observations suggest that running short-run LD may be fundamentally different from obtaining samples from the EBMs, and therefore the maximum likelihood explanation for the training procedure may be invalid. 

\citet{nijkamp2019learning} first study the intriguing properties of short-run LD. They conjecture that the short-run LD behaves more like a generator model, and try to explain the maximum likelihood training by introducing $q_{\theta}$, the marginal distribution of the sample after running $K$ steps of LD starting from a fixed initial distribution. However, they do not study $q_{\theta}$ with an explicit formulation. In this paper, we follow their work to provide an alternative understanding of the maximum likelihood training of EBMs. In particular, we replace the LD sampling with noise-free dynamics, so that the output samples are produced by a deterministic transformation of the initial points. In this case, we regard the dynamic as a generator model, and the initial points as latent variables. By ensuring that the generator is invertible, we can explicitly study the density of the distribution induced by the sampling dynamics (where the randomness is entirely determined by the initial points). In addition, by treating the sampling dynamics as a generator model, we find that we can improve the sample quality by adding the generator loss term from GANs to the original loss.

\section{Noise-free sampling dynamics as flow model}\label{noise free section}
In this section, we demonstrate how to explicitly obtain the density induced by the noise-free sampling dynamics by enforcing invertibility. We start by replacing the Langevin dynamics in (\ref{LD_cont}) with the noise-free  gradient descent ODE:
\begin{align}\label{ODE}
\rvx^{\prime}(t)=-\nabla_\rvx E_\theta(\rvx(t)), \quad
\rvx(0)=\rvx_{0}, \quad t \in [0, T],
\end{align}
which is guaranteed to produce an invertible map under very mild conditions on $E$, and we can write the continuous flow
 \citep{chen2018neural,grathwohl2018ffjord}: 
\begin{align}\label{ode solve}
    \rvx_T = G^T_\theta(\rvx_0) = \operatorname{ODESolve}(-\nabla_\rvx E_\theta (\rvx(t)), \rvx_0, [0, T]).
\end{align}
Since there is no noise term, given $\rvx_0$, the process can be represented by a deterministic generator model with latent variable $\rvx_0$. We want to emphasize that $T$ is an important component of the generator model, and we should use roughly the same T when sampling. Moreover, as $G^T_\theta(\rvx_0)$ is invertible, the likelihood along the path can be obtained by instantaneous change of variables formula \citep{chen2018neural}, and
the log likelihood of data $\rvx$ under the flow model can be computed by 
\begin{align}\label{ode_lik}
    \log p(\rvx) = \log p(\rvx_0) + \int_{0}^{T} \text{tr}\left[\nabla_{\rvx\rvx}E_\theta(\rvx(t))\right] dt.
\end{align}

As a special case, the forward Euler solver for this equation yields:
\begin{align}\label{NFLD formula}
    \rvx_{t+1}=\rvx_{t}-\frac{\eta}{2} \nabla_{\rvx} E_\theta(\rvx_t),\quad t = 0,1, \cdots,K-1,
\end{align}
with initialization $\rvx_0$ from some fixed simple distribution $p_0$ in $\mathbb{R}^{d}$ such as the standard Gaussian. 
In particular, $G_\theta(\rvx_0): \mathbb{R}^{d} \rightarrow \mathbb{R}^{d}$ is a residual flow \citep{pmlr-v97-behrmann19a}:
\begin{align}
    \rvx_K = G_\theta(\rvx_0) = (I -  \frac{\eta}{2} \nabla_{\rvx} E_\theta)^K (\rvx_0).
\end{align}
$G_\theta(\rvx_0)$ is guaranteed to be invertible if $\operatorname{lip}\left(\frac{\eta}{2} \nabla_{\rvx} E_\theta)\right)<1$ \citep{pmlr-v97-behrmann19a}. This holds as long as $\nabla_{\rvx} E_\theta$ has bounded Lipschitz constant and the step size $\eta$ is sufficiently small. However, it is still difficult to choose the step size that ensures invertibility, and therefore, we generalize $G_\theta(\rvx_0)$ to be any numerical solution to the initial value ODE problem (\ref{ODE}).

To summarize, we train the energy network $E_{\theta}$ by doing the gradient update (\ref{derivative}) with negative samples obtained from (\ref{ode solve}). After training, we can obtain new samples by running (\ref{ode solve}), and compute the likelihood of data point $\rvx$ by solving the ODE (\ref{ODE}) in the \textit{reverse} direction to find the corresponding initial point $\rvx_0$ and then apply (\ref{ode_lik}). 


\section{Connection with W-GAN and the generator loss term} \label{gan connection section}
It is well known that the maximum likelihood training of EBMs is closely related to the training of Wasserstein-GANs \citep{arjovsky2017wasserstein}, where the objective for the discriminator $D$ (assuming $D$ is $1$-Lipschitz) is 
\begin{align}\label{wgan}
\max _{D} \underset{\rvx \sim p_D}{\mathbb{E}}\left[D(\rvx)\right]-\underset{\tilde{\rvx} \sim p_{G}}{\mathbb{E}}\left[D(\tilde{\rvx})\right].
\end{align}
The gradient of (\ref{wgan}) is (up to a sign) very similar to (\ref{derivative}) except that here the negative samples are drawn from the generator, while in (\ref{derivative}), the negative samples are drawn from the EBM itself. Note that the sign does not matter as we can model the negative energy instead. W-GANs use the discriminator $D$ to contrast true data and samples generated by the generator $G$, while EBMs use the energy function $E$ to contrast true data and samples generated by $E$ itself implicitly through MCMC. Therefore, the maximum likelihood training of EBMs can be described as a \textit{self-adversarial game}. 

In W-GAN, after the discriminator is updated by (\ref{wgan}), the generator $G$ is then updated by
\begin{align}
    \max _{G} \underset{\tilde{\rvx} \sim p_{G}}{\mathbb{E}}\left[D(\tilde{\rvx})\right].
\end{align}
In other words, it maximizes the discriminator's output for fake samples generated by $G$. Strictly speaking, there is no corresponding loss term in the training of EBMs, as the sampling is done by MCMC rather than deterministic mapping. However, as discussed in section \ref{noise free section}, in practice the sampling process can be seen as a generator model with initial points as latent variables. In this case, we actually have an \textit{explicit} generator $G_{\theta}$ defined in (\ref{ode solve}), and therefore we can update the parameter of $G_{\theta}$ (which is just $\theta$) by the following objective:
\begin{align}
\min _{\theta}   E_{\text{sg}(\theta)}(G_{\theta}(\rvx_0)),
\end{align}
where $\rvx_0$ is the latent variables sampled from $p_0$, and sg($\cdot$) is the stop gradient operation. Here we stop the gradient of $E_{\theta}$ because we only want to differentiate through the generation process. 

Hence, we propose to add the extra update step for $G_{\theta}$ at each iteration, so that we are essentially training a W-GAN whose discriminator and generator share the same set of parameters, and  conjecture that the adversarial training will improve the sample quality. 

One modification is made for the implementation. Typically when training GANs, we alternate the update of the parameters of the discriminator and the generator and hence two batches of samples are generated. This can be slow in our case as drawing samples requires iterative updates. Therefore, we use the same batch of samples to update $E_{\theta}$ and $G_{\theta}$, and since we only have one set of parameters $\theta$, it is equivalent to optimizing the following objective:
\begin{align}\label{gan loss total}
    \min _{\theta}E_{\theta}(\rvx) - E_{\theta}(G_{\text{sg}(\theta)}(\rvx_0))+ E_{\text{sg}(\theta)}(G_{\theta}(\rvx_0)), \quad \rvx \sim p_D,\rvx_0 \sim p_0.
\end{align}
We will empirically study the effectiveness of this training objective in section \ref{image exp}.

\section{Related Work} 
Our work is closely related to earlier studies on the properties of ML training EBMs with short-run non-convergent MCMC \citep{nijkamp2019learning,nijkamp2020anatomy}, where they illustrate through experiments that the short-run LD behaves more like a generator model, and in particular \citet{nijkamp2019learning} provide a moment matching framework for explaining the mechanism behind the maximum likelihood training. In addition, \citet{xie2018cooperative,xie2020learning} propose MCMC teaching, where a separate generator is trained to absorb the process of LD sampling. This suggests that their method is based on the assumption that LD used in practice can be represented as a generator model. Additionally, \citet{han2019divergence} provides a probabilistic way to deal with EBM without MCMC. We take a further step from them to explicitly study the properties of the generator models.

Since our noise-free sampling dynamics can yield an invertible gradient flow , our work is related to the concept of generative gradient flows \citep{zhang2018monge,huang2021convex}. In addition, \citet{song2021scorebased} show that the stochastic dynamics of score based generative models \citep{song2019generative, ho2020denoising} are equivalent to specific deterministic ODE flows \cite{chen2018neural, grathwohl2018ffjord}. However, such equivalence cannot be easily established for Langevin diffusion. \cite{pang2020learning} connects EBM and generator model, but what they do is learning an EBM prior for the generator. Finally, our work is related to  previous work that connects GANs with EBMs \citep{che2020your, song2020discriminator,ansari2021refining} or invertible flows \citep{grover2018flow}. In particular, \citet{grover2018flow} use invertible structure, such as real-NVP \citep{dinh2016density}, for the generator of GANs, but they focus on hybrid training with adversarial and maximum likelihood objectives. 

\section{Results}\label{results}
In this section, we conduct experiments to verify our proposed methods and arguments in section \ref{noise free section} and \ref{gan connection section}. Specifically, we train energy functions on 2d-toy data and image data by replacing the MCMC sampling with deterministic dynamics. Throughout the experiments, we initialize the dynamics with noise sampled from standard Gaussian distribution. We do not use persistent sampling, as we want to interpret the model as a generator with fixed prior. The deterministic dynamics can be simply defined by (\ref{NFLD formula}), or more generally the path to solve the ODE as in (\ref{ode solve}). In particular, we need to use the latter method if we want to compute the density induced by the dynamics. More experimental detail can be found in appendix \ref{experiment detail}.

\begin{figure}[t]
    \centering
    \begin{subfigure}{.9\linewidth}
    \includegraphics[scale=0.44]{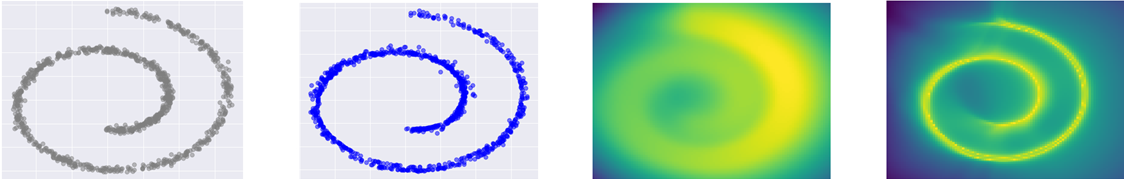}
        \caption{Swiss roll}
    \end{subfigure}
    \begin{subfigure}{.9\linewidth}
    \includegraphics[scale=0.44]{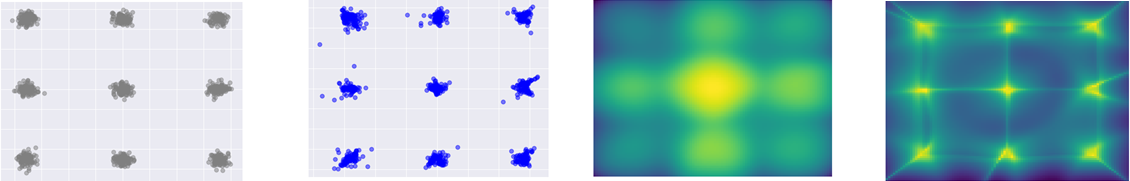}
       \caption{Gaussian grid}
    \end{subfigure}
    \caption{\label{toy plot}
      For each toy dataset, \textbf{column 1:} samples from the true data distribution; \textbf{column 2:} samples from the ODE flow; \textbf{column 3:} (unnormalized) log density of the EBM by plotting the value of $-E_{\theta}(\rvx)$; \textbf{column 4:} log density of the ODE flow computed by (\ref{ode_lik}). The spurious connections between components will visually disappear if we take exponential (see Appendix \ref{section 2d toy additional}). We plot log density because the sampling dynamics directly use it.}
\end{figure}

\subsection{2D toy data}
We use the Swiss roll and 9 Gaussian mixture grid as the true distributions, and our energy function $E_{\theta}: \mathbb{R}^{2} \rightarrow \mathbb{R}$ is a simple neural network with several fully connected layers. We use the neural ODE formulation and solve the ODE in (\ref{ODE}) with the default Dormand–Prince solver as in \citet{chen2018neural}. In Figure \ref{toy plot}, we plot the samples obtained from solving the ODE (\ref{ode solve}). We also plot the log density of the ODE flow and the value of the negative energy function (which is the unnormalized log density of the corresponding EBM) in the same figure. We observe that we can obtain good samples, even though the densities of the EBMs are not close to the ground truth densities. In contrast, the density functions induced by the ODE flow captures the densities the true data distributions very well. We also train EBMs with valid MCMC sampling with noise term, and plot the density functions and generated samples in appendix \ref{section 2d toy additional}. There we make a similar observation that the densities of EBMs do not match the data distribution.

These observations provide evidence that  maximum likelihood training of EBMs is actually training a gradient flow model. Since the density defined by the final energy function completely fails to capture the true data density, arguments that running the sampling dynamics draws samples from the EBM is certainly incorrect; instead, we show that the dynamic \textit{itself} is the generative model to sample from, as its density matches the shape of the true density.

In addition, we also train the ODE flows with the same formulation and structure using the maximum likelihood objective (where the likelihood is defined in (\ref{ode_lik})) and compare the obtained data likelihood with that of the flows trained by the EBM objective. For the ODE flows trained by maximum likelihood, the test data log likelihood (averaged over 10000 test samples) is \textbf{-0.69} nats on Swiss roll and \textbf{-1.47} nats on Gaussian grid. The test data likelihood of the ODE flows trained by the EBM objective is \textbf{-0.86} nats and \textbf{-1.95} nats on these two datasets, respectively. As expected, the flows directly trained by maximizing data likelihood have higher test likelihood, but the flows trained by the EBM objective still perform reasonably well. 

\subsection{Image data}\label{image exp}
Experiments on toy data reveal that the maximum likelihood training of EBMs may actually lead to training generator or flow models. If this is true, then the noise term used in Langevin dynamics may be unnecessary or even harmful. We confirm this by studying the sample quality on common image datasets including MNIST, Fashion-MNIST, CIFAR-10 and CelebA. For simplicity, our energy functions are simple convolutional nets instead of more complex residual networks used in \citet{du2019implicit,xiao2021vaebm}, and therefore we only compare relative performances. 

\begin{figure}[t]
    \centering
    \begin{subfigure}{.9\linewidth}
    \includegraphics[scale=0.44]{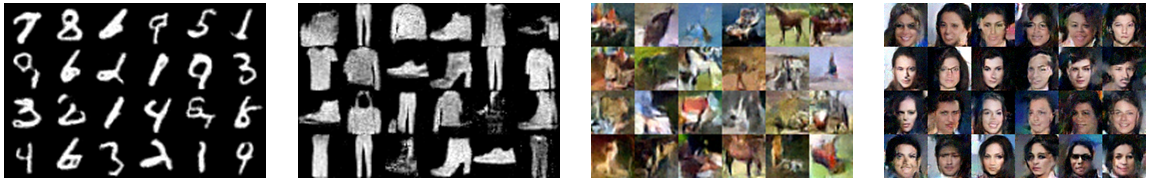}
        \caption{EBMs w/ noisy dynamics}
    \end{subfigure}
    \begin{subfigure}{.9\linewidth}
    \includegraphics[scale=0.44]{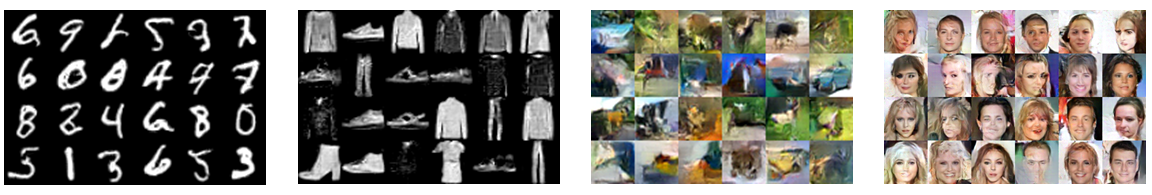}
        \caption{EBMs w/ noise-free dynamics}
    \end{subfigure}
    \begin{subfigure}{.9\linewidth}
    \includegraphics[scale=0.44]{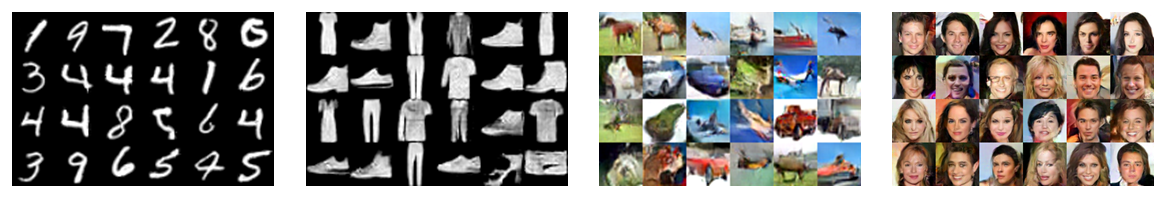}
       \caption{EBMs w/ noise-free dynamics + generator loss}
    \end{subfigure}
    \caption{\label{qualitative}
     Qualitative samples of models with noisy or noise-free sampling dynamics, and models with extra loss to update the generator defined by the dynamics.}
\end{figure}

We train energy functions using the gradient update in (\ref{derivative}), where the samples are generated by either noisy or noise-free sampling dynamics. For noise-free dynamics, we use the gradient descent formulation in (\ref{NFLD formula}) instead of the neural ODE formulation, because we only want to study the effect of noise while keeping all other factors the same. For noisy sampling dynamics, we reduce the noise scale as done in almost all other work, otherwise the training diverges quickly. We report the FID scores \citep{heusel2017gans} in Table \ref{fid table}, and qualitative samples in Figure \ref{qualitative} and appendix \ref{additional qualitative}. We observe that EBMs trained with noise-free dynamics indeed obtain better sample quality on all datasets. Besides, we plot the loss curves in Appendix \ref{loss curve}, and we find that removing the noise significantly improves the training stability. Even with reduced noise scale, training EBMs with noisy sampling dynamics may still diverges during training. These results suggest that the noise term in sampling dynamics may have negative effects, which further supports the argument that the we should treat the model as a generator defined by the gradient of the energy instead of an EBM. 

\begin{table}[t]
\centering
\caption{FID scores on image datasets for different models}

\begin{tabular}{lcccc}\label{fid table}
 & MNIST & Fashion-MNIST & CIFAR-10 & CelebA
\\ \hline
EBM w/ noisy dynamic & 15.4& 61.7 & 70.4 & 69.6\\
EBM w/ noise-free dynamic & 11.7 & 50.1& 61.6 & 56.6  \\
\shortstack{\\ + Generator loss}
& 7.7 & 40.6 & 47.9 & 34.8\\
\hline
\end{tabular}
\end{table}

Treating the noise-free dynamics as generator models, we further apply the additional adversarial loss term for the W-GAN generator update discussed in section \ref{gan connection section}. In particular, we train the model with loss in (\ref{gan loss total}). We report the FID in the last line of Table \ref{fid table}, where we find the generator update significantly improves the sample quality. Qualitative samples are shown in Figure \ref{qualitative} and additional samples in appendix \ref{additional qualitative}. This experiment shows that the noise-free dynamics is indeed a generator, and we can use it to train GANs. 

\section{Conclusion and Discussion}
In this paper, we provide new insights to understand the maximum likelihood training of EBMs. We believe that instead of training EBMs, the maximum likelihood objective actually trains generator models through a self-adversarial mechanism. The generator model is defined implicitly by the gradient of the energy network, and we study the property of the generator model by removing the noise in the MCMC sampling dynamics. We conduct experiments to justify our thoughts and make the following observations:
\begin{itemize}
    \item On toy data, the density function induced by the invertible noise-free dynamics is close to the shape of the true data density, while the density of the EBM with corresponding energy function fails to capture the true density.
    \item On image datasets, we observe that removing the noise in the LD improves sample quality and training stability.
    \item The sample quality can be further improved by introducing the generator update discussed in section \ref{gan connection section}, i.e., making the self-adversarial game into an adversarial game.
\end{itemize}
These observations together suggest that the mechanism behind the ML training of EBMs is to train a generator or gradient flow model, and we can benefit from removing the noise in the sampling dynamics. As a result, given the difficulty of running MCMC in high dimensions, we should study the convergence of MCMC sampling in high dimensions more carefully, and probably focus more on training techniques without sampling, if our goal is to train valid energy-based models.

\newpage
\bibliography{iclr2021_conference}
\bibliographystyle{iclr2021_conference}

\newpage
\appendix
\section{Experimental settings} \label{experiment detail}
In this section, we introduce detailed settings of our experiments.
\subsection{2D toy data}
On 2D toy data, we use a 5-layer fully connected networks with 256 hidden units and swish activation function. We train our models with Adam optimizer, with constant learning rate $1e-3$. The models are trained for 3000 iterations with batch size 800. 

We draw negative samples by solving the ODE in (\ref{ode solve}). To do so, we use the solver implemented by \citet{chen2018neural}. We set the initial value to random samples from 2-d standard Gaussian distribution. We use the default dopri5 solver, $T \in [0,0.2]$, and numerical error tolerance tolerance $1e-5$. After training, samples are drawn by solving the same neural ODE.

\subsection{Image data}
We resize MNIST and Fashion-MNIST to $32 \times 32$. The network structures are presented in Table \ref{netstruct}.
We train all models with Adam optimizer with learning rate $5e-4$ and batch size $64$. As we mention in the main text, the training of EBMs with noisy dynamics is unstable and it will diverge after certain number of iterations. This is also observed in \citet{du2019implicit} and \citet{xiao2021vaebm}. Therefore, we follow their setting to train the EBMs until divergence. For EBMs trained with noise-free dynamics, we found the training to be more stable. We set the number of training iterations similar to that of EBMs with noisy dynamics. In particular, we train 8000 iterations for MNIST and Fashion-MNIST, 40000 iterations for CIFAR-10 and 30000 iterations for CelebA. To draw negative samples, we set the step size to be $0.1$ and number of steps to be $40$ for MNIST/Fashion-MNIST and $60$ for CIFAR-10 and CelebA. For the noisy sampling dynamics, we set the noise scale to be $0.1$.

For the extra GAN loss, we need to store the gradient while running the gradient descent steps (\ref{NFLD formula}). This can be done by setting the $\text{create graph}$ option when computing the gradient in PyTorch's auto differential package \citep{paszke2019pytorch}.

\begin{table}
  \centering
  \begin{tabular}{ccc}
    \toprule
    MNIST   & CIFAR-10 & CelebA    \\
    \midrule
     $3 \times 3$ $\text{Conv}_{\text{nf}}$ Stride $1$  & $3 \times 3$ $\text{Conv}_{\text{nf}}$ Stride $1$  & $3 \times 3$ $\text{Conv}_{\text{nf}}$ Stride $1$\\
     
     $4 \times 4$ $\text{Conv}_{2 \times \text{nf}}$ Stride $2$ & $4 \times 4$ $\text{Conv}_{2 \times \text{nf}}$ Stride $2$ & $4 \times 4$ $\text{Conv}_{2 \times \text{nf}}$ Stride $2$\\
     
     $4 \times 4$ $\text{Conv}_{2 \times \text{nf}}$ Stride $2$ & $4 \times 4$ $\text{Conv}_{4 \times \text{nf}}$ Stride $2$ & $4 \times 4$ $\text{Conv}_{4 \times \text{nf}}$ Stride $2$ \\
     
     $4 \times 4$ $\text{Conv}_{2 \times \text{nf}}$ Stride $2$&  $4 \times 4$ $\text{Conv}_{8 \times \text{nf}}$ Stride $2$& $4 \times 4$ $\text{Conv}_{8 \times \text{nf}}$ Stride $2$\\
     Faltten, FC layer to scalar & Faltten, FC layer to scalar& $4 \times 4$ $\text{Conv}_{16 \times \text{nf}}$ Stride $2$\\
     &&Faltten, FC layer to scalar\\
    \bottomrule
  \end{tabular}
   \caption{Network structures for different datasets. nf means number of filters. For MNIST, Fashion MNIST and CelebA, nf = 32; for CIFAR-10, nf = 64. Swish activation is applied after each convolutional layer. }\label{netstruct}
\end{table} 

\section{Additional results on 2D toy data} \label{section 2d toy additional}
\begin{figure}[t]
    \centering
    \begin{subfigure}{.9\linewidth}
    \includegraphics[scale=0.5]{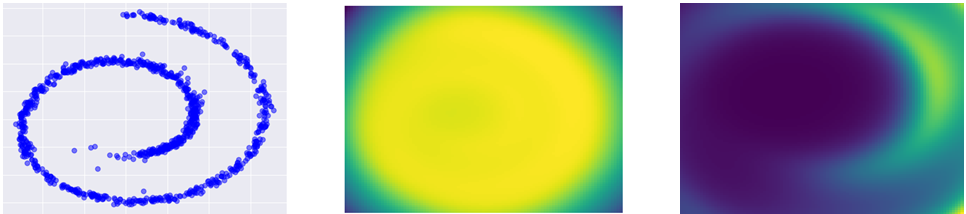}
        \caption{Swiss roll}
    \end{subfigure}
    \begin{subfigure}{.9\linewidth}
    \includegraphics[scale=0.5]{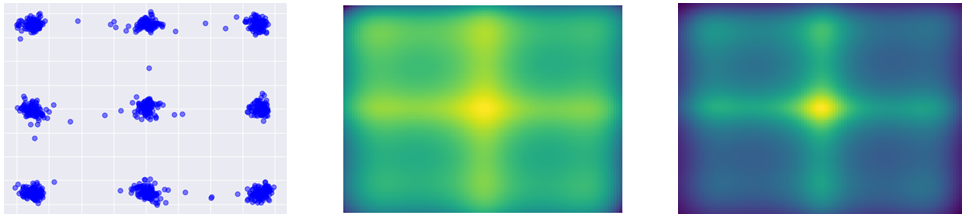}
       \caption{Gaussian grid}
    \end{subfigure}
    \caption{\label{toy plot app}
     Results of EBMs trained and sampled from using noisy dynamics on toy data. For each sub-figure, we plot the \textbf{left:} samples obtained from running Langevin dynamics, \textbf{middle:} (unnormalized) log density of the EBM , and \textbf{right:} normalized density of the EBM, where the normalization constant is estimated by numerical integration.}
\end{figure}

In Figure \ref{toy plot app}, we plot the samples and (unnormalized) density of EBMs trained with noisy sampling dynamics (finite steps of Langevin dynamics). The shown samples are generated by running the same Langevin dynamics. Note that we cannot plot the density induced by the dynamic itself as the noise term makes it not invertible. However, we have similar observations as in Figure \ref{toy plot}. While the sample quality is good, the density of EBMs completely fail to match the true data distribution. We notice that \citet{gao2021learning} make similar observations (see their Figure 2).

This suggest that even in the usual case where EBMs are trained and sampled with LD, samples are actually generated from a (noise injected) generator model instead of the EBM. Therefore, the mechanism behind maximum likelihood training of EBMs is actually training generator models implicitly defined by the sampling dynamics.

\begin{figure}[t]
    \centering
    \begin{subfigure}{.9\linewidth}
    \includegraphics[scale=0.6]{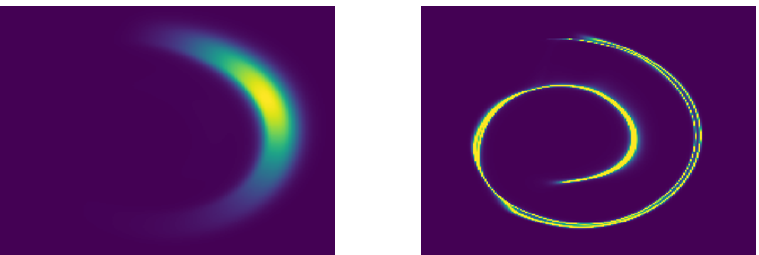}
        \caption{Swiss roll}
    \end{subfigure}
    \begin{subfigure}{.9\linewidth}
    \includegraphics[scale=0.6]{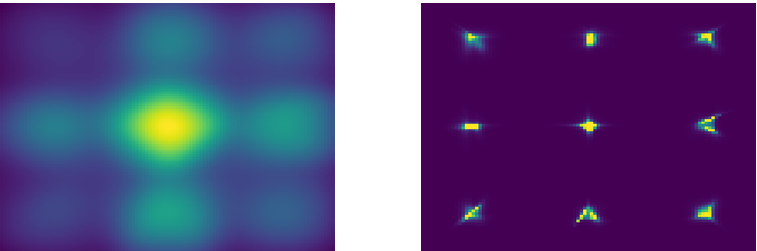}
       \caption{Gaussian grid}
    \end{subfigure}
    \caption{\label{normalized density}
      For each sub-figure, \textbf{left:} normalized density of the EBM, and \textbf{right:} density of the gradient flow.}
\end{figure}

We also plot the normalized density of the EBMs and gradient flows in Figure \ref{normalized density}, where we observe that the spurious high density region shown in the log density plot in Figure \ref{toy plot}  disappears, and we still find that the density of the gradient flows captures the true density much better than that of the EBMs.

\section{Loss curves} \label{loss curve}
\begin{figure}[t]
    \centering
    \begin{subfigure}{.49\linewidth}
    \includegraphics[scale=0.4]{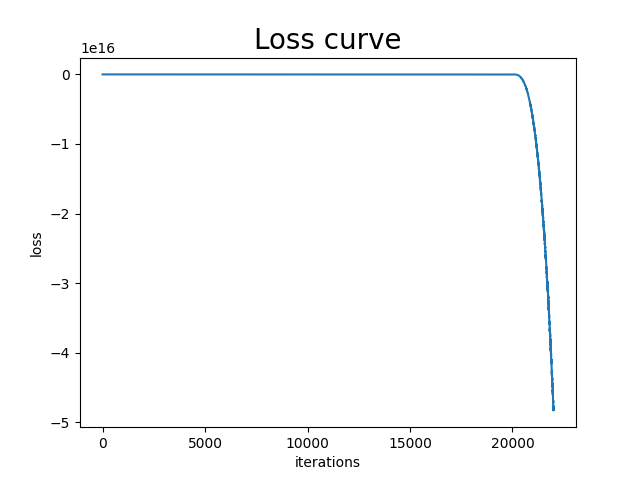}
        \caption{EBMs w/ noisy dynamics}
    \end{subfigure}
    \begin{subfigure}{.49\linewidth}
    \includegraphics[scale=0.4]{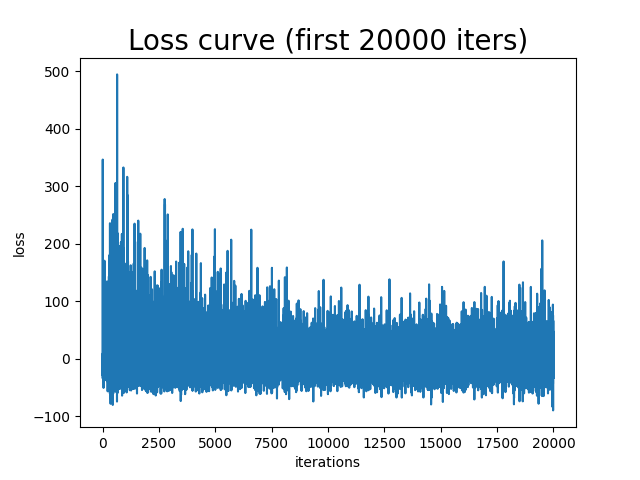}
        \caption{EBMs w/ noisy dynamics, first 20000 iterations}
    \end{subfigure}
    \begin{subfigure}{.49\linewidth}
    \includegraphics[scale=0.4]{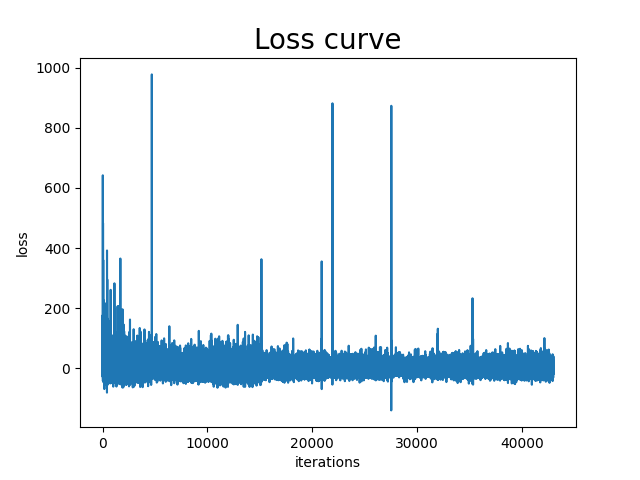}
       \caption{EBMs w/ noise-free dynamics}
    \end{subfigure}
    \begin{subfigure}{.49\linewidth}
    \includegraphics[scale=0.4]{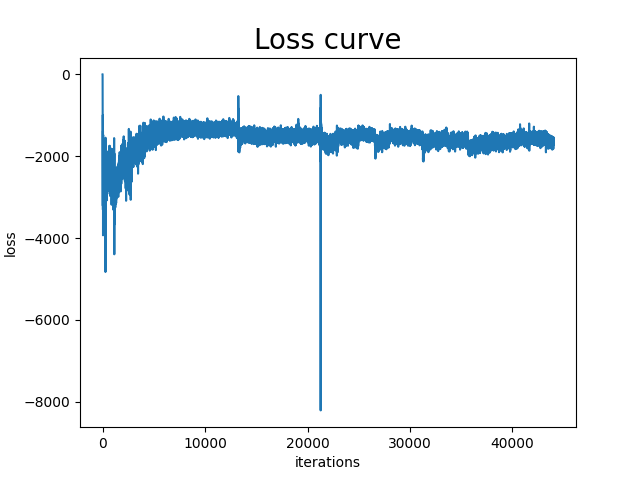}
       \caption{EBMs w/ noise-free dynamics + generator loss}
    \end{subfigure}
    \caption{\label{loss plot}
     Plots of loss curves on CIFAR-10 dataset. \textbf{(a):} When sampling using the noisy MCMC, the training diverges after 20000 iterations. \textbf{(b):} For better visualization, we plot the loss curve for the first 20000 iterations. \textbf{(c):} When using noise-free dynamics, the training is more stable. \textbf{(d):} With the additional generator loss, although we see some jumps on the loss curve, the training is overall stable.}
\end{figure}

In Figure \ref{loss plot}, we plot the loss curve along the training of models with noisy or noise-free dynamics on CIFAR-10. We observe that for both models, the losses oscillate around zero, as observed in \citet{nijkamp2020anatomy}. However, the model trained with noisy dynamics diverges after 20000 iterations, while the the training of model with noise-free dynamics is much more stable. In addition, we observe that adding the extra generator loss as discussed in section \ref{gan connection section} does not affect the training stability.

\section{Additional results on image data}\label{additional qualitative}
\begin{figure}[t]
    \includegraphics[scale=0.7]{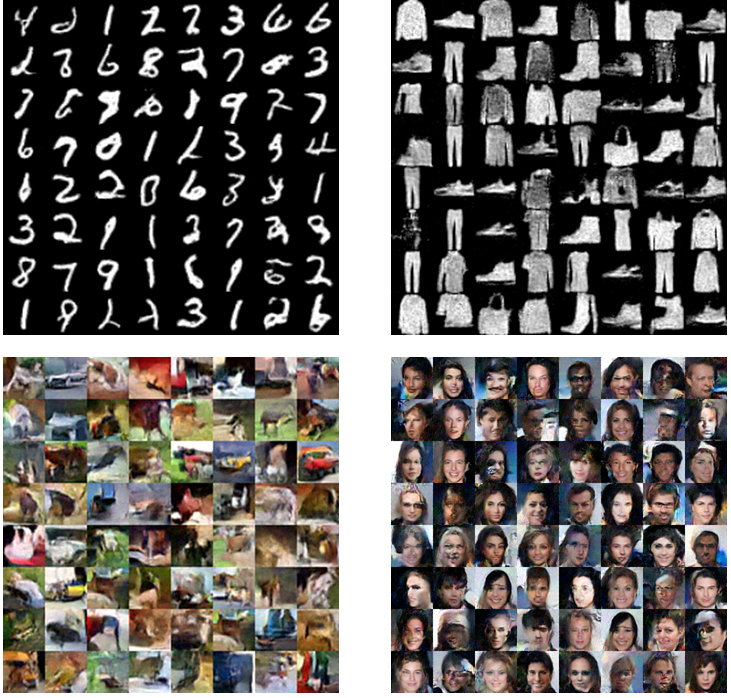}
        \caption{\label{4a}Additional samples from EBMs w/ noisy dynamics}
\end{figure}

\begin{figure}[t]
    \includegraphics[scale=0.7]{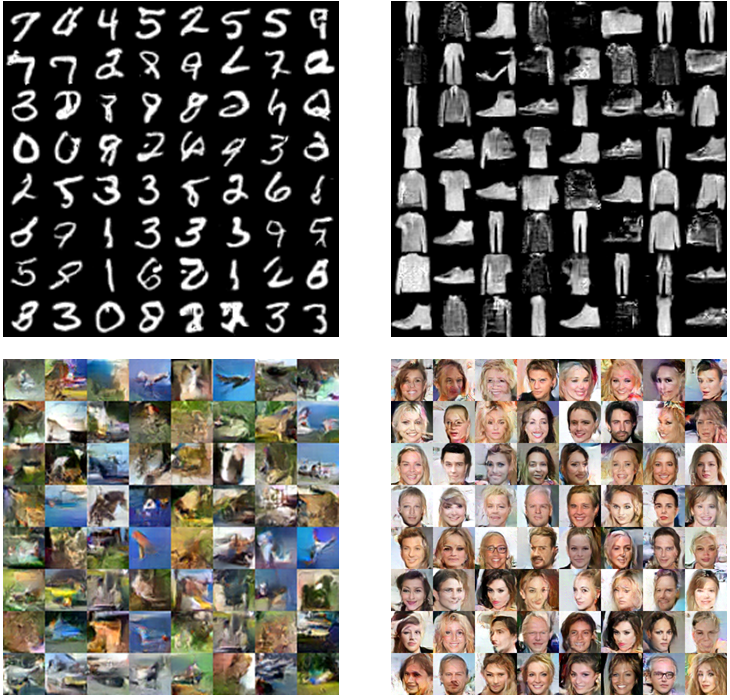}
        \caption{\label{4b}Additional samples from EBMs w/ noise-free dynamics}
\end{figure}

\begin{figure}[t]
    \includegraphics[scale=0.7]{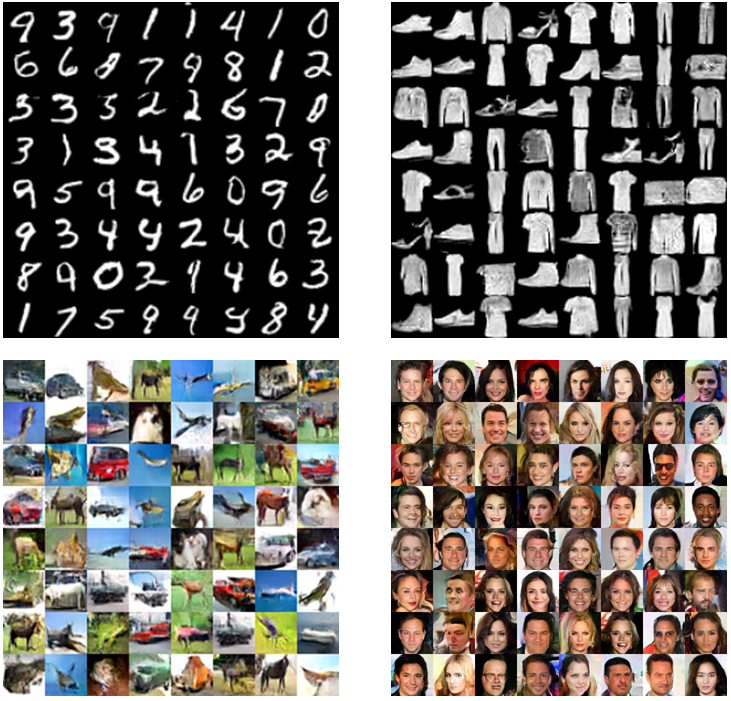}
        \caption{\label{4c}Additional samples from EBMs w/ noise-free dynamics plus extra generator loss}
\end{figure}

In Figure \ref{4a}, \ref{4b} and \ref{4c}, we present additional qualitative samples corresponding to Figure \ref{qualitative} in the main text.

\end{document}